\documentclass[twoside,leqno,twocolumn]{article}
\usepackage{times}  
\usepackage{helvet}  
\usepackage{courier}  
\usepackage{url}  
\usepackage{graphicx}  
\usepackage{tikz}
\usepackage{amsopn}
\usepackage{bbding}
\usepackage{amsmath}
\usepackage{amsfonts} 
\usepackage{amsmath,amssymb}
\usepackage{bbm}
\usepackage{algorithm}
\usepackage{amssymb}
\usepackage{caption}
\usepackage[noend]{algpseudocode}
\usepackage[utf8]{inputenc} 
\usepackage[T1]{fontenc}    
\usepackage{amsmath}
\usepackage{amsfonts} 
\usepackage{subcaption}
\usepackage{epstopdf}
\usetikzlibrary{calc,trees,positioning,arrows,chains,shapes.geometric,%
	decorations.pathreplacing,decorations.pathmorphing,shapes,%
	matrix,shapes.symbols,shapes.gates.logic.US,trees,positioning,arrows}

\newlength\myindent
\newtheorem{theorem}{Theorem}[section]

\newtheorem{lemma}[theorem]{Lemma}

\begin{document}

\tikzset{%
	brace/.style = { decorate, decoration={brace, amplitude=5pt} },
	mbrace/.style = { decorate, decoration={brace, amplitude=5pt, mirror} },
	label/.style = { black, midway, scale=0.5, align=center },
	toplabel/.style = { label, above=.5em, anchor=south },
	leftlabel/.style = { label,rotate=-90,left=.5em,anchor=north },   
	bottomlabel/.style = { label, below=.5em, anchor=north },
	force/.style = {scale=0.9 },
	round/.style = { rounded corners=2mm },
	legend/.style = { right,scale=0.4 },
	nosep/.style = { inner sep=0pt },
	generation/.style = { anchor=base }
}
\tikzstyle{data}=[draw, fill=white!20, text width=1.7em, 
text centered, minimum height=2.5em, rounded corners,]
\tikzstyle{n}=[draw, fill=white!20, text width=3.4em, 
text centered, minimum height=2.3em]
\tikzstyle{ann} = [above, text width=5em, text centered]
\tikzstyle{wa} = [sensor, text width=10em, fill=white!20, 
minimum height=6em, rounded corners, drop shadow]
\tikzstyle{sc} = [sensor, text width=13em, fill=white!20, 
minimum height=10em, rounded corners, drop shadow]

\title{\Large Mining the Unstructured: Accelerate RNN-based Training with Importance Sampling}
\author{Fei Wang\thanks{Department of Computer Science and Engineering, Shanghai Jiao Tong University. bomber@sjtu.edu.cn, \{gao-xf, gchen\}@cs.sjtu.edu.cn} \\
	\and
	Xiaofeng Gao$^{\star}$ \\
	\and
	Weichen Li\thanks{Facebook, Inc. weichenli@fb.com} \\
	\and
	Guihai Chen$^{\star}$ \\
	\and
	Jun Ye\thanks{Intel Asia-Pacific R\&D Ltd. jason.y.ye@intel.com}}
\date{}

\maketitle







\begin{abstract} \small\baselineskip=9pt 
	Importance sampling (IS) as an elegant and efficient variance reduction (VR) technique for the acceleration of stochastic optimization problems has attracted many researches recently. Unlike commonly adopted stochastic uniform sampling in stochastic optimizations, IS-integrated algorithms sample training data at each iteration with respect to a weighted sampling probability distribution $P$, which is constructed according to the precomputed importance factors. Previous experimental results show that IS has achieved remarkable progresses in the acceleration of training convergence. Unfortunately, the calculation of the sampling probability distribution $P$ causes a major limitation of IS: it requires the input data to be well-structured, i.e., the feature vector is properly defined. Consequently, recurrent neural networks (RNN) as a popular learning algorithm is not able to enjoy the benefits of IS due to the fact that its raw input data, i.e., the training sequences, are often unstructured which makes calculation of $P$ impossible. In considering of the the popularity of RNN-based learning applications and their relative long training time, we are interested in accelerating them through IS. This paper propose a novel Fast-Importance-Mining algorithm to calculate the importance factor for unstructured data which makes the application of IS in RNN-based applications possible. Our experimental evaluation on popular open-source RNN-based learning applications validate the effectiveness of IS in improving the convergence rate of RNNs.
\end{abstract}

\section{Introduction}
For the optimization of general finite-sum problems, e.g., empirical risk minimization (ERM), stochastic gradient descent (SGD) may be the most widely adopted optimizer algorithm. Assume $\phi_{i}$, $\forall i\in\{1,2,...,N\}$ are vector functions that map $\mathbb{R}^d\to\mathbb{R}$, $i\sim P$ means $i$ is drawn iteratively with respect to sampling probability distribution $P$. This paper studies the following optimization problem:
\begin{equation}
\label{opt}
\mathop{\mathrm{min}}_{w\in \mathbb{R}^d}F(w):=\mathbb{E}_{i\sim P}(\phi_i(w)+\eta r(w))
\end{equation}
where $r(w)$ is the regularizer and $\eta$ is the regularization factor. Denote $f_i(w)=\phi_i(w)+\eta r(w)$, for stochastic gradient descent optimization, $w$ is updated as:
\begin{equation}
\label{update}
w_{t+1} = w_t-\lambda\nabla f_{i_t}(w_t)
\end{equation} 
where ${i_t}$ is the index sampled at $t$-th iteration and $\lambda$ is the step-size. The key advantage of SGD is that the computation of the stochastic gradient is much faster than the true gradient. It is easy to notice that with a proper sampling distribution $P$ (typically a uniform distribution), the stochastic gradient equals to the true gradient in expectation. Assume $p_i^t$ as the probability of selecting $x_i$ as the training sample at iteration $t$, we denote by
\begin{equation}
\mathbb{V}\big((np^t_{i_t})^{-1}\nabla f_{i_t}(w_t)\big)=\mathbb{E}\|(np^t_{i_t})^{-1}\nabla f_{i_t}(w_t)-\nabla F(w_t)\|_2^2
\end{equation}
the gradient variance. It is commonly known that the convergence procedure is severely slowed down when gradient variance is large.

To cope with this problem, original importance sampling (IS) algorithms \cite{alain2015variance, Siddharth, hazan2011beating, Bach} are developed for variance reduction in stochastic optimizations by adjusting the sampling probability distribution $P$ iteratively. Despite of its theoretical effectiveness, in industrial applications such algorithms are seldom adopted for their iteratively re-estimation of the sampling probability is computationally infeasible. 

Recently, optimized IS algorithms inspired by randomized Kaczmarz algorithm were proposed by authors of \cite{csiba2016importance, Needell, Strohmer, p_zhao}. This kind of IS algorithms sample training data at each iteration $t$ w.r.t to a pre-constructed probability distribution instead of iterative re-estimation of $P$. The sampling distribution $P$ is constructed based on the supremem of the gradient norm of the data samples, i.e., $\sup \|\nabla f_i(w)\|_2$. When cooperated with a proper step-size, the complexity of the upper-bound of training iterations to achieve certain accuracy is proved to be decreased. In the rest of this paper, we call this kind of gradient-norm-supremum based importance sampling algorithm directly as IS for clarity. With the success of IS, many stochastic optimizations have adopted it and show impressive convergence results improvements, e.g., SVM-based classification etc.

On the other hand, with the fast development of neural network (NN)s, they have been widely used in modeling many large-scale optimization problems and have demonstrated great advantages. Among all these research fields, recurrent neural networks (RNN) have unique capabilities in modeling context-conditioned sequences such as natural language processing (NLP), acoustic sequences processing, time series prediction and achieves much better training performance (accuracy, convergence rate) than previous state-of-the-art. 

Since we consider IS as an elegant VR technique and RNN a widely used learning algorithm, we are very interested in deploying IS in RNN applications which yields both practical significance and novelty. On the other hand, as is mentioned above, the sampling probability $p_i$ is based on the supremum of $\|\nabla f_i(w)\|_2$, which in turn involves the calculation of $\|x_i\|_2$. This poses a major problem of the current IS algorithms, i.e., the calculation of the sampling distribution $P$ requires the input data to be well-structured, i.e., the feature vector of the data samples are properly defined, otherwise $\|x_i\|_2$ can not be calculated. Unfortunately, for RNN applications such as NLP, its raw data are typically randomly mapped into a $d$-dimensional space which is in fact unstructured. In fact, in this kind of RNN-based optimization problems, the feature vector is only structured by multiplying a mapping matrix $W$, which is also to be learned during the optimization, e.g., the embed matrix $W^{emb}$ for the embedding (structuring) of the input word vector $x_i$ in RNN-based NLP. As a consequence, the sampling probability distribution $P=\{p_i\},\forall i\in\{1,2,...,N\}$ can not be constructed. 

To solve this bottlenecking problem, we propose a novel algorithm which mines the importance of RNN training samples w.r.t which the sampling probability $p_i$ can be calculated and thus the IS can be proceeded in RNN-based applications, i.e., IS-RNN. Our experimental results validate that our IS-RNN algorithm is able to decrease the gradient variance and achieves better convergence results than conventional non-IS RNN.

This paper is organized as follows. In section 2 we give brief descriptions about necessary preliminaries and concepts of the IS algorithm. We discuss the difficulty of applying current IS in RNN in Section 3. In section 4, we propose and analyze our optimized IS algorithm for RNN in detail. The evaluation results of IS on some most popular RNN-based applications for convergence acceleration are shown in Section 5. In the last section, we make conclusion of this paper.

\section{Importance Sampling for Stochastic Optimizations}
We first briefly introduce some key concepts of IS. Like most previous related literatures, we make the following necessary assumptions for the convergence analysis of the target stochastic optimization problems.
\begin{itemize}
	\item $f_i$ is strongly convex with parameter $\mu$, that is:
	\begin{equation}
	\label{convexity}
	\langle x-y,\nabla f_i(x)-\nabla f_i(y) \rangle \ge \mu \|x-y\|_2^2, \, \forall x,y \in \mathbb{R}^d
	\end{equation}
	\item Each $f_i$ is continuously differentiable and $\nabla f_i$ has Lipschitz constant $L_i$ w.r.t $\|\cdot\|_{\star}$, i.e.,
	\begin{equation}
	\label{continuity}
	\|\nabla f_i(x)-\nabla f_i(y)\|_{\star}\le L_i\|x-y\|_{\star}, \, \forall x, y \in \mathbb{R}^d
	\end{equation}
\end{itemize}
Where $\forall i \in \{1,2,...,N\}$.
\subsection{Importance Sampling}
Importance sampling reduces the gradient variance through a non-uniform sampling procedure instead of drawing sample uniformly. For conventional stochastic optimization algorithms, the sampling probability of $i$-th sample at $t$-th iteration, i.e., $p_i^t$, always equal to $1/N$ while in an IS scheme, $p_i^t$ is endowed with an importance factor $I_i^t$ and thus the $i$-th sample is sampled at $t$-th iteration with a weighted probability:
\begin{equation}
p^{t}_i=I^t_i/N, \quad s.t. \sum_{i=1}^{N}p_i^t=1
\end{equation}
where $N$ is the number of training samples. With this non-uniform sampling procedure, to obtain an unbiased expectation, the update of $w$ is modified as:
\begin{equation}
\label{up}
w_{t+1} = w_t-\frac{\lambda}{np^t_{i_t}}\nabla f_{i_t}(w_t)
\end{equation}
where $i_t$ is drawn i.i.d w.r.t the weighted sampling probability distribution $P^t=\{p_i^t\}, \forall i\in \{1,2,...,N\}$.
\subsection{Importance Sampling for Variance Reduction} Recall the optimization problem in Equation \ref{opt}, using the analysis result from \cite{p_zhao}, we have the following lemma:
\begin{lemma}
	Set $\sigma^2=\mathbb{E}\|\nabla f_i(w_{\star})\|_2^2$ where $w_{\star}=\arg\underset{w}{\min}F(w)$. Let $\lambda \le \frac{1}{\mu}$, with the update scheme defined in Equation \ref{up}, the following inequality satisfy:
	\begin{equation}
	\begin{aligned}
	\mathbb{E}[F(w_{t+1})&-F(w_{\star})]\le \frac{1}{2\lambda}\mathbb{E}[\|w_{\star}-w_t\|_2^2-\|w_{\star}-w_{t+1}\|_2^2]\\ 
	&-\mu\mathbb{E}\|w_{\star}-w_t\|_2^2+\frac{\lambda_t}{\mu}\mathbb{E}\mathbb{V}\Big((np_{i_t}^t)^{-1}\nabla f_{i_t}(w_t)\Big)
	\end{aligned}
	\end{equation}
	where the variance is defined as $\mathbb{V}\big((np^t_{i_t})^{-1}\nabla f_{i_t}(w_t)\big)=\mathbb{E}\|(np^t_{i_t})^{-1}\nabla f_{i_t}(w_t)-\nabla F(w_t)\|_2^2$, and the expectation is estimated w.r.t distribution $P^t$.
\end{lemma}
In order to minimize the gradient variance, it is easy to verify that the optimal sampling probability $p_i^t$ is:
\begin{equation}
\label{precise}
p_i^t=\frac{\|\nabla f_{i}(w_t)\|_2}{\sum_{j=1}^N\|\nabla f_{j}(w_t)\|_2}, \qquad \forall i \in \{1,2,...,N\}.
\end{equation}
Obviously, such iteratively re-estimation of $P^t$ is completely impractical. The authors propose to use the supremum of $\|\nabla f_{i}(w_t)\|_2$ as an approximation. Since we have L-Lipschitz of $\nabla f_i$, assume $\|w_t\| \le R$ for any $t$, we get $\|\nabla f_{i}(w_t)\|_2 \le RL_i$, i.e., $\sup\|\nabla f_{i}(w_t)\|_2=RL_i$. Thus the actual sampling probability $p_i$ is calculated according to:
\begin{equation}
\label{actual}
p_i=\frac{L_i}{\sum_{j=1}^{N}L_j}, \, \forall i \in\left\{1,2,...,N\right\}.
\end{equation} 
\begin{algorithm}[t]
	\caption{Practical Importance Sampling for SGD}
	\begin{algorithmic}[1]
		\Procedure{IS-SGD}{$T$}
		\State Construct Sampling Distribution $P$ According to Equation \ref{actual}
		\State Generate Sample Sequence $S$ w.r.t distribution $P$.
		\State Sample $i_t$ from $\{i\}_{i=0}^n$ w.r.t distribution $P$.
		\For{$i=0;i\not=T;i$++}
		\State $i_t=S[i]$
		\State $w_{t+1} = w_t-\frac{\lambda}{np_{i_t}}\nabla f_{i_t}(w_t)$
		\EndFor\label{IS-practical}
		\EndProcedure
	\end{algorithmic}
\end{algorithm}
The authors prove that IS accelerated SGD achieves a convergence bound as:
\begin{equation}
\frac{1}{T}\sum_{t=1}^{T}\mathbb{E}[F(w_t)-F(w_{\star})]\le \sqrt{\frac{\|w_{\star}-w_0\|_2^2}{\sigma}}\left(\frac{\sum_{i=1}^{n}L_i}{n}\right)\frac{1}{T},
\end{equation}
while for standard non-IS SGD optimizers that actually samples $x_i$ w.r.t uniform distribution, the convergence bound is:
\begin{equation}
\frac{1}{T}\sum_{t=1}^{T}\mathbb{E}[F(w_t)-F(w_{\star})]\le \sqrt{\frac{\|w_{\star}-w_0\|_2^2\sum_{i=1}^{n}(L_i^2)}{\sigma n}} \frac{1}{T},
\end{equation}
when $\lambda$ is set as $\sqrt{\sigma\|w_{\star}-w_0\|_2^2}/\left(\frac{\sum_{i=1}^{n}L_i}{n}\sqrt{T}\right)$. According to Cauchy-Schwarz inequality, we always have:
\begin{equation}
\frac{n\sum_{i=1}^{n}L_i^2}{(\sum_{i=1}^{n}L_i)^2} \ge 1,
\end{equation}
which implies that IS does improve convergence bound and the improvements get more significant when $\frac{(\sum_{i=1}^{n}L_i)^2}{\sum_{i=1}^{n}(L_i^2)}\ll n$. 

Clearly, $p_i$ of each $f_i$ can be analyzed beforehand. For example, for L2-regularized optimization SVM problem with squared hinge loss, i.e., 
\begin{equation}
f_i(w)=(\lfloor 1-y_iw^{T}x_i \rfloor_{+})^2+\frac{\lambda}{2}\|w\|_2^2
\end{equation}
where $x_i$ is the $i$-th sample and $y_i \in \{-1,+1\} $ is the corresponding label, $\|\nabla f_{i}(w)\|_{2}$ can be bounded as 
\begin{equation}
\|\nabla f_{i}(w)\|_{2} \le 2(1+\|x_{i}\|_{2}/\sqrt{\lambda})\|x_{i}\|_{2}+\sqrt{\lambda}
\end{equation}
Since $\|x_i\|_2$ is the only variable in the calculation of $p_i$, the sampling distribution $P$ can be constructed off-line completely. The pseudo code of practical IS-SGD algorithm can be written as Algorithm 1. As can be seen that, the core procedure of IS is the construction of $P$. Once $P$ is constructed, IS-SGD works as same as SGD except that the training sample is selected w.r.t to P and the step-size is adjusted with ${1}/{p_i}$. This means that IS as an effective VR technique can be implemented with almost no extra online computation which makes it very suitable for VR of large scale optimization problems.

\section{Recurrent Neural Network with IS}
Recurrent neural networks \cite{funahashi1993approximation, mikolov2010recurrent, pineda1987generalization, wang1993analysis} have been developed with great efforts due to their strong advantages in modeling large-scale sequences. However, comparing to convolutional neural networks (CNN), its training is typically much difficult and slower. 

As we have discussed previously, IS has been recently studied in many optimization problems such as SVM-based linear-regression or even CNN applications \cite{Alain}. However we found that the research of applying IS in RNN is still left missing. In considering of the popularity and the novelty, we are motivated to study the application of IS in RNN. Our goal is to improve the convergence results of RNN training through IS not only effectively but also efficiently. To discuss our algorithm, we first give a brief introduction of RNN since some important concepts are used in designing our optimized IS algorithm for RNNs.
\begin{figure}[t]
	\label{rnn-arch}
	\centering
	\caption{Unrolled Architecture of RNN with 3 Iterations}
	\begin{tikzpicture}[x=2.4cm, y=2.0cm]
	\fill [draw,fill=none,rounded corners] (-1.3,-0.7) rectangle (-1.0,-0.0);
	\filldraw (-1.15,-0.55) circle (3pt);
	\filldraw (-1.15,-0.35) circle (3pt);
	\filldraw (-1.15,-0.15) circle (3pt);
	
	\fill [draw,fill=none,rounded corners] (-0.5,-0.7) rectangle (-0.2,-0);
	\filldraw (-0.35,0-0.55) circle (3pt);
	\filldraw (-0.35,0-0.35) circle (3pt);
	\filldraw (-0.35,0-0.15) circle (3pt);
	
	\fill [draw,fill=none,rounded corners] (0.3,-0.7) rectangle (0.6,-0);
	\filldraw (0.45,0-0.55) circle (3pt);
	\filldraw (0.45,0-0.35) circle (3pt);
	\filldraw (0.45,0-0.15) circle (3pt);
	
	\fill [draw,fill=none,rounded corners] (-1.7,-1.3) rectangle (-1.3,-1.0);
	\filldraw (-1.5,-1.15) circle (3pt);
	\fill [draw,fill=none,rounded corners] (-0.9,-1.3) rectangle (-0.5,-1.0);
	\filldraw (-0.7,-1.15) circle (3pt);
	\fill [draw,fill=none,rounded corners] (-0.1,-1.3) rectangle (0.3,-1.0);
	\filldraw (0.1,-1.15) circle (3pt);
	
	\draw [line width=0.4mm,->] (-1.8,-0.35)   -- (-1.3,-0.35);
	\draw [line width=0.4mm,->] (-1.0,-0.35)   -- (-0.5,-0.35);
	\draw [line width=0.4mm,->] (-0.2,-0.35)   -- (0.3,-0.35);
	\draw [line width=0.4mm,->] (0.6,-0.35)   -- (1.2,-0.35);
	
	\draw [line width=0.4mm,->] (-1.15,0)   -- (-1.15,0.35);
	\draw [line width=0.4mm,->] (-0.35,0)   -- (-0.35,0.35);
	\draw [line width=0.4mm,->] (0.45,0)   -- (0.45,0.35);
	
	\draw [line width=0.4mm,->,to path={|- (\tikztotarget)}] (-1.5,-1) to (-1.3,-0.55);
	\draw [line width=0.4mm,->,to path={|- (\tikztotarget)}] (-0.7,-1) to (-0.5,-0.55);
	\draw [line width=0.4mm,->,to path={|- (\tikztotarget)}] (0.1,-1) to (0.3,-0.55);
	
	\node[text width=2cm] at (-0.9,0.5) {$y_{t}$};
	\node[text width=2cm] at (-1.1,-0.1) {$W$};
	\node[text width=2cm] at (-1.55,-0.25) {$h_{t-1}$};
	\node[text width=2cm] at (-1.55,-0.91) {$x_{i_{t}}$};
	
	\node[text width=2cm] at (-0.05,0.5) {$y_{t+1}$};
	\node[text width=2cm] at (-0.27,-0.1) {$W$};
	\node[text width=2cm] at (-0.45,-0.25) {$h_{t}$};
	\node[text width=2cm] at (-0.7,-0.91) {$x_{i_{t+1}}$};
	
	\node[text width=2cm] at (0.8,0.5) {$y_{t+2}$};
	\node[text width=2cm] at (0.55,-0.1) {$W$};
	\node[text width=2cm] at (0.3,-0.25) {$h_{t+1}$};
	\node[text width=2cm] at (0.1,-0.91) {$x_{i_{t+2}}$};	
	
	\node[text width=2cm] at (1.1,-0.25) {$h_{t+2}$};
	\end{tikzpicture}
\end{figure}
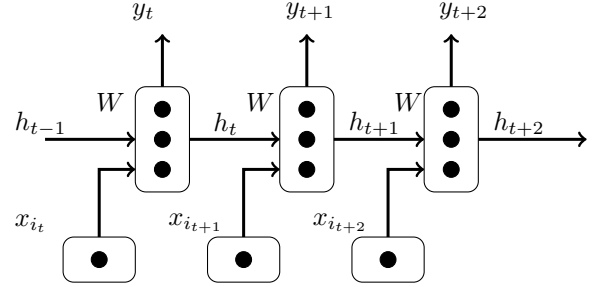
\subsection{Recurrent Neural Network}
RNNs are proposed for sequence modeling. Its variants e.g., long short term memory (LSTM) and gated recurrent units (GRU) have achieved significant progresses in machine learning tasks such as NLP and acoustic sequences analyze due to its capability in conditioning the model on all previous inputs. Figure 1 shows the general architecture of RNN. For the $i$-th raw training sample input sequence, i.e., $X_i=\{X_{i}^1,X_{i}^2,...,X_{i}^n\}$ where $n$ is the size of the training sample $X_i$. RNN embeds $X_i$ to $x_i$ and reads in one element of the input vector, i.e., $x_{i}^t$ at each iteration $t$ and calculates the prediction $y_t$ and hidden state $h_t$. The calculation step is shown in the following Equation:
\begin{equation}
\begin{aligned}
\label{rnn}
\centering
&x_i=W^{emb}X_i\\
&h_t=\sigma(W^{h}h_{t-1}+W^{x}x_{i}^t+b^h)\\
&y_t=softmax(W^{s}h_t+b^y)
\end{aligned}
\end{equation} 
\begin{itemize}
\item $W^{emb} \in \mathbb{R}^{N_v\times d}$: embedding matrix used to convert raw word $x_{raw}$ into embedded value, to be learned.
\item $x_{i}^t \in \mathbb{R}^d$: the $t$-th element of embedded training sample $x_i$.
\item $W^{x} \in \mathbb{R}^{D_h\times d}$: weight matrix used to condition the input $x_t$, to be learned.
\item $W^{h} \in \mathbb{R}^{D_h\times D_h}$: weight matrix used to condition the previous hidden state $h_{t-1}$, to be learned.
\item $W^s \in \mathbb{R}^{N_v\times D_h}$: weight matrix used to performed the classification $h_t$, to be learned.
\item $h_{t-1} \in \mathbb{R}^{D_h}$: hidden state output of last iteration.
\item $\sigma$: non-linear activation functions.
\item $b^h \in \mathbb{R}^{D_h}$, $b_y \in \mathbb{R}^N_v$: bias terms.
\end{itemize}
where $d$ is the size of embedding, $D_h$ is the number of hidden neurons and $N_v$ is the size of vocabulary. With the corresponding output vector of $x_i$ denoted as $y_i=\{y_{i_1}, y_{i_2},...,y_{i_n}\}$ and $y'_i=\{y'_{i_1},y'_{i_2},...,y'_{i_n}\}$ as the true probability distribution. The objective function is calculated as multiclass cross-entropy, i.e.,:
\begin{equation}
	\mathcal{L}_i^T(y_i,y'_i)=\frac{1}{T}\sum_{t=1}^{n}y'_{i_t}\log(y_{i_t})
\end{equation}
\subsection{Problems of Applying IS in RNN}
During the research of applying IS in RNN to accelerate its training procedure, we met two main bottlenecking problems caused by the special architecture of RNN. We give detailed analysis as the following.
\subsubsection{Base-Gradient}
Recall that in IS-SGD, the sampling probability $p_i$ of training data $x_i$ is approximated by the supremum of its gradient norm $\sup \|\nabla f_i(w_t)\|_2$. The first question is that in RNN based on what gradient the sampling distribution is computed. For example, in a simple linear-regression problem $y=W^Tx+b$, we have only one model (linear regression matrix $W$) to be trained and thus we have no ambiguousness in $\sup \|\nabla f_i(w_t)\|_2$. However in RNN, it can be seen from equation \ref{rnn} that we have four models $W^{emb}$, $W^h$, $W^x$ and $W^s$ to be learned (assume $b^h$ and $b^y$ are not considered) and consequently 4 corresponding gradients, i.e., $g^{emb}_i=\frac{\partial \mathcal{L}_i}{\partial W^{emb}}$, $g^{h}_i=\frac{\partial \mathcal{L}_i}{\partial W^{h}}$, $g^{x}_i=\frac{\mathcal{L}_i}{\partial W^{x}}$ and $g^{s}_i=\frac{\mathcal{L}_i}{\partial W^{s}}$. 

Intuitively we have two choices, the first is to use the combination of the four gradients, i.e., $G(g^{emb}_i,g^{h}_i,g^{x}_i,g^{s}_i)$ where $G$ is the combination function. Another way is to choose one of the four gradients based on which its norm supremum is used to calculate $p_i$ according to Equation \ref{actual}. While we have no guidance in designing the combination function $G$, the second choice is simpler and most importantly, more easy to explain, e.g., we may base the calculation of $P$ on $g^{emb}$ if we consider the update of $W^{emb}$ affects the convergence more significantly than others. For clarity, we denote by $\nabla^{base}_i$ the gradient based on which $p_i$ is calculated, i.e., the base-gradient.
\subsubsection{Unstructured Input}
The second bottlenecking problem is that once the form $\nabla^{base}$ is selected, obtaining its norm supremum $\sup \|\nabla^{base}_i\|_2$ is very difficult due to the fact that $W^{emb}$ is not learned at the beginning which means that the embedded value of the input $x_i$, based on which the all four gradients are actually computed, is not decided. In other words, the feature vector of input $x_i$ is not properly structured until $W^{emb}$ is converged. With the learning of $W^{emb}$, the structuring of the feature vector keeps proceeding and the value of $x_i$ varies accordingly which brings in much uncertainty in the analyzing of $\sup \|\nabla^{base}_i\|_2$.

In brief, the special architecture of RNN makes the attempt to accelerates its convergence through IS very difficult. To enjoy the benefits of IS, a special modified variant of IS algorithm that solves the mentioned two bottlenecking problems efficiently for RNN is needed.
\section{Mining the Importance for RNN}
We propose an optimized IS algorithm which can be applied in RNN to achieve accelerated convergence procedure by targeting at the two above mentioned bottlenecking problems.
\subsection{Base-Gradient Selection}
Our first step is to select a proper $\nabla^{base}$. As we have discussed above, we select $\nabla^{base}_i$ as one of the four gradients, i.e., $g^{emb}_i,g^{h}_i,g^{x}_i,g^{s}_i$ for simplicity and easy explanation. We first look into the detailed derivation of the four gradients:
\begin{equation}
g_i^h =\sum_{t=1}^{T}\sum_{k=1}^{t}\frac{\partial \mathcal{L}_i^t}{\partial y_{t}}\frac{\partial y_t}{\partial h_{t}}\left(\prod_{j=k+1}^{t}\frac{\partial h_j}{\partial h_{j-1}}\right)\frac{\partial h_k}{\partial W^{h}}
\label{s}
\end{equation}

\begin{equation}
g_i^x=\sum_{t=1}^{T}\frac{\partial \mathcal{L}_i^t}{\partial y_{t}}\frac{\partial y_t}{\partial h_{t}}\frac{\partial h_t}{\partial W^{x}}
\end{equation}

\begin{equation}
g_i^s=\sum_{t=1}^{T}\frac{\partial \mathcal{L}_i^t}{\partial y_{t}}\frac{\partial y_t}{\partial W^{s}}
\end{equation}

\begin{equation}
g_i^{emb}=\sum_{t=1}^{T}\frac{\partial \mathcal{L}_i^t}{\partial y_{t}}\frac{\partial y_t}{\partial h_{t}}\frac{\partial h_t}{\partial x_{i_t}}\frac{\partial x_{i_t}}{\partial W^{emb}}
\label{e}
\end{equation}
A reasonable option is to select $\nabla^{base}$ as the gradient which has the fundamental effect for the whole training procedure. Obviously, according to Equation \ref{rnn} and Equation \ref{s} to \ref{e}, $x_i$ is the only input conditioned by $W^x$ and $h_t$ is an intermediate memory value that is actually decided by $h_0$ and the training sequence $<x_i,y'_i>, \forall i \in \{1,2,...,T\}$. In fact, it is easy to verify that with determined initialization models, i.e., $b^h_0$, $b^y_0$, $W^h_0$, $W^s_0$ and $h_0$, all gradients are totally decided by the sequence of $\{W^xx_i,y_i'\}, i \in\{1,2,...,t\}$. We thus empirically choose $W^x$ as our base-gradient $\nabla_{base}$\footnote{In fact, we also tested other gradients during the experimental evaluations. The result verifies our analysis, i.e., choosing $g_i^x$ as the $\nabla_{base}$ based on which $P$ is constructed makes IS in RNN effective while others are either non-effective or less effective.}, that is, $p_i$ is calculated as:
\begin{equation}
	p_i=\frac{\sup\|g_i^x\|_2}{\sum_{j=1}^{N}\sup\|g_j^x\|_2}, \qquad \forall i \in \{1,2,...,N\}.
\end{equation}
After the selection of $\nabla^{base}_i=g_i^x$, we are now left to obtain, or construct an appropriate proxy of $\sup\|\nabla^{base}_i\|_2$.
\subsection{Mining the Importance}
Since we have selected the base-gradient, obtaining or finding proxy of $\sup \|\nabla^{base}_i\|_2$ is the following step. As we have discussed above, $x_i=W^{emb}X_{i}$ varies with the proceeding of the training which makes the bounding of $\|\nabla^{base}_i\|_2$ technically difficult.

Intuitively, one possible method is to periodically re-calculate $P$ according to the snapshot of $W^{emb}$. That is, denote by $W^{emb}_s$ the snapshot of $W^{emb}$ at iteration $s$, calculate embedded dataset $x_i^s=\{W^{emb}_{s}X_i\}, \forall i\in\{1,2,...,N\}$ and estimate $p_i^{s}$ according to $x_i^s$ respectively. Indeed, it seems that with the structuring of the feature vector (i.e., the training of $W^{emb}$) getting more accurate, the corresponding $p_i^s$ is getting closer to the optimum. However, this procedure is based on an important assumption, which is, the initial sampling distribution $P^0=\{p_i^0\}, \forall i \in \{1,2,...,N\}$ calculated based on $x_i^0=W^{emb}_0X_i$ is better than the uniform distribution. Otherwise, the training procedure is more likely to be slowed down due to an inferior initial sampling distribution $P^0$. Unfortunately, as is obvious, this requirement of the initially-structured feature vector is in fact not guaranteed.  

In considering the fact that obtaining exact $\sup \|\nabla^{base}_i\|_2$ is complex and difficult, we here propose another simple yet effective Fast-Importance-Mining (FIM) algorithm for the construction of proxy of $\sup \|\nabla^{base}_i\|_2$. See Figure 2 for illustration, the FIM algorithm trains each data sample in parallel to obtain the proxy value which serves as the same function with $\|\nabla^{base}_i\|_2$. Each $x_i$ has a \textbf{private} trained $RNN_i$ with itself as the \text{only} training sample. The training for each $RNN_i$ starts with the same initial value of RNN and ends with the same accuracy, $\epsilon$, i.e., $\mathcal{L}_i\le \epsilon$. Empirically, we set $\epsilon$ much smaller (typically two magnitude lower) than the standard training for RNN, i.e., trained with the whole dataset. The accuracy is easy to be met since it has only one training sample. Denote by $W^x_i$ the matrix $W^x$ in corresponding $RNN_i$, with the trained $W^x$ set $W=\{W^x_i\}, \forall i \in \{1,2,...,N\}$, we use $\|W^x_i\|_2$ as the proxy value of $\sup \|\nabla^{base}_i\|_2$ and thus the sampling probability $P=\{p_i\},\forall i \in \{1,2,...N\}$ is calculated as:
\begin{equation}
\label{pi}
	p_i=\frac{\|W^x_i\|_2}{\sum_{j=1}^{N}\|W^x_j\|_2}, \qquad \forall i \in \{1,2,...,N\}.
\end{equation}
\begin{figure}[t]
	\label{IS-fig}
	\centering
	\caption{Fast Importance Mining}
	\begin{tikzpicture}[x=2.4cm, y=2.0cm]
	
	\node[text width=2cm] at (0.35,1.4) {$x_1$};
	\node[text width=2cm] at (0.15,0.8) {$RNN_1$};
	\draw [line width=0.2mm,->] (-0.05,1.3)  -- (-0.05,1);
	\fill [draw,fill=none] (-0.3,0.6) rectangle (0.2,1);
	\draw [line width=0.2mm,->] (-0.05,0.6)  -- (-0.05,0.3);
	\draw[thick, ->] (0.25,0.85) arc (170:-170:0.3cm);
	\node[text width=2cm] at (0.3,0.15) {$\|W^x_1\|_2$};

	\node[text width=2cm] at (1.25,1.4) {$x_2$};
	\node[text width=2cm] at (1.05,0.8) {$RNN_2$};
	\draw [line width=0.2mm,->] (0.85,1.3)  -- (0.85,1);
	\fill [draw,fill=none] (0.6,0.6) rectangle (1.1,1);
	\draw [line width=0.2mm,->] (0.85,0.6)  -- (0.85,0.3);
	\draw[thick, ->] (1.15,0.85) arc (170:-170:0.3cm);
	\node[text width=2cm] at (1.2,0.15) {$\|W^x_2\|_2$};
	\node[text width=2cm] at (2.05,1.4) {$\boldsymbol{\cdot}\boldsymbol{\cdot}\boldsymbol{\cdot}$};
	\node[text width=2cm] at (2.05,0.8) {$\boldsymbol{\cdot}\boldsymbol{\cdot}\boldsymbol{\cdot}$};
	\node[text width=2cm] at (2.05,0.1) {$\boldsymbol{\cdot}\boldsymbol{\cdot}\boldsymbol{\cdot}$};
	\node[text width=2cm] at (2.05,-1.4) {$\boldsymbol{\cdot}\boldsymbol{\cdot}\boldsymbol{\cdot}$};
	
	\node[text width=2cm] at (2.75,1.4) {$x_N$};
	\node[text width=2cm] at (2.55,0.8) {$RNN_N$};
	\draw [line width=0.2mm,->] (2.35,1.3)  -- (2.35,1);
	\fill [draw,fill=none] (2.1,0.6) rectangle (2.6,1);
	\draw [line width=0.2mm,->] (2.35,0.6)  -- (2.35,0.3);
	\draw[thick, ->] (2.65,0.85) arc (170:-170:0.3cm);
	\node[text width=2cm] at (2.7,0.15) {$\|W^x_N\|_2$};
	
	\draw [line width=0.2mm,->] (0.1,0.04)  -- (1.15,-0.45);
	\draw [line width=0.2mm,->] (1.,0.04)  -- (1.25,-0.4);
	\draw [line width=0.2mm,->] (2.45,0.04)  -- (1.45,-0.45);
	\draw (1.3,-0.6) circle (0.4cm);
	\node[text width=2cm] at (1.65,-0.6) {$\sum$};
	
	\draw [line width=0.2mm,->] (1.15,-0.8)  -- (0.1,-1.15);
	\draw [line width=0.2mm,->] (1.25,-0.83)  -- (1.,-1.15);
	\draw [line width=0.2mm,->] (1.45,-0.8)  -- (2.45,-1.15);	
	\node[text width=3cm] at (0.3,-1.4) {$p_1=\frac{\|W^x_1\|_2}{\sum_{j=1}^{N}}$};
	\node[text width=2cm] at (1.1,-1.4) {$p_2=\frac{\|W^x_2\|_2}{\sum_{j=1}^{N}}$};
	\node[text width=2cm] at (2.6,-1.4) {$p_N=\frac{\|W^x_N\|_2}{\sum_{j=1}^{N}}$};
	\end{tikzpicture}
\end{figure}
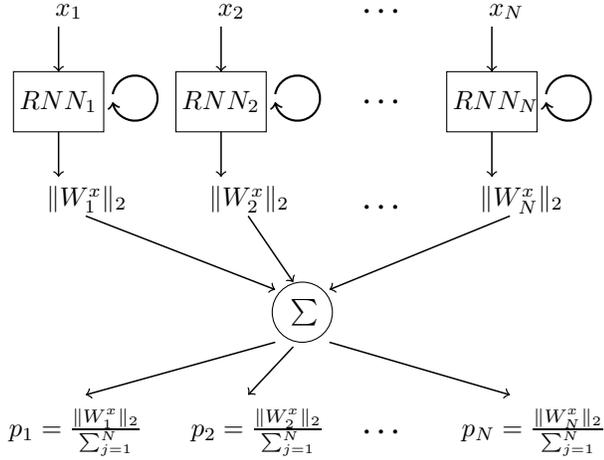
\begin{algorithm}[t]
	\caption{Fast Importance Mining}
	\begin{algorithmic}[1]
		\Procedure{Construct\_Sampling\_Distribution}{$\epsilon$}
		\State \textbf{Parallel} do for $\forall i \in \{1,2,...,N\}$:
		\State \qquad $RNN_i$=\textbf{new} RNN($W(0),b(0)$)
		\State \qquad $\mathcal{L}_i =$ MAX\_FLOAT
		\State \qquad \textbf{While} $\mathcal{L}_i\ge \epsilon$
		\State \qquad \qquad $\mathcal{L}_i$=Train\_$RNN_i$$(x_i$)
		\State $W_{sum} = \sum_{j=1}^{N}\|W^x_j\|_2$
		\State \textbf{Parallel} do for $\forall i \in \{1,2,...,N\}$:
		\State \qquad $p_i=\frac{\|W^x_i\|_2}{W_{sum}}$
		\EndProcedure
	\end{algorithmic}
\label{xx}
\end{algorithm}
\subsection{Analysis}
In considering the fact that $p_i$ is calculated in the form of $\frac{a_i}{\sum_{j=1}^{N}a_j}$ rather an absolute value directly, it is more easy to find a proxy value that faithfully reflects the sampling importance of $x_i$ rather than approximate the exact value of $\sup \|g_i^x\|_2$ which is technically difficult. 

According to our proposed FIM algorithm, $W^x_i$ is actually the sum of all history gradients and initial value $W^x(0)$, i.e.,
\begin{equation}
W^x_i=W^x(0)-\sum_{t=1}^{T_i}g_i^x(t)
\end{equation}
where we denote by $g_i^x(t)$ as $g_i^x$ at iteration $t$ and $T_i$ the iterations used for $RNN_i$ to be trained with accuracy $\epsilon$. Due to the triangle inequality, we further have:
\begin{equation}
	\|W^x_i\|_2\le\|W^x(0)\|_2 + \sum_{t=1}^{T_i}\|g_i^x(t)\|_2
\end{equation}
It is reasonable to conclude that data samples $x_i$ with larger $\|W_i^x\|_2$ is to have larger $\|g_i^x\|_2$ which complies with our expectation of a proxy value and our intuition of giving more sampling importance to $x_i$ with larger $\|g_i^x\|_2$. 
 
We still face the unstructured feature vector problem in FIM, however it is reasonable to say that once $RNN_i$ is trained, the feature vector of $x_i$ is structured (at least for itself). Consequently, since each $RNN_i, \forall i \in \{1,2,...,N\}$ is trained separately with its only training sample $x_i$, the corresponding structuring rules that have been learned, i.e., $W^{emb}_i, \forall i \in \{1,2,...,N\}$, are different from each other and the global rule $W^{emb}$ trained with the whole dataset. Such differences make $X_i$ embedded into different $x_i$ when in whole dataset training and FIM. However according to our statistical evaluation, the difference between $W^{emb}_i, \forall i \in \{1,2,...,N\}$ and $W^{emb}$ is not significant and thus we empirically assume that such inconsistency will not hurt the performance of IS heavily.

Meanwhile, since the structuring of $x_i$ is proceeded with the training process, it is reasonable to condition the whole structuring procedure as a method to avoid the effect of structuring variance during the FIM. Thus we consider $W^x_i=W^x(0)-\sum_{t=1}^{T_i}g_i^x(t)$ and use its norm $\|W_i^x\|_2$ instead of just recording $\max \|g_i^x(t)\|_2$ as the proxy value of $\sup\|g_i^x(t)\|_2$. The FIM algorithm of $P$ is rather empirical, however it works surprisingly well according to our evaluation as to be shown in next section.
\section{Experimental Results}
To validate the effectiveness of IS in RNN, we conduct experimental evaluations\footnote{Our testbed is a server with Intel Xeon E5-2699 v4 CPU and 128G DDR3 RAM.} on two popular RNN-based machine learning tasks: LSTM for sentence classification and RNN-RBM for polyphonic music sequence modeling. The source code we based on is from the well-developed deep learning tutorials of Theano\cite{bergstra2010theano} which is popularly used in academic deep learning researches. Particularly, since the networks of our target evaluation tasks are variants of standard RNN thus the implementation of IS changes accordingly as we will describe in detail for each case. All source code of our experimental evaluation along with the visualization script can be accessed from the author's git repository\footnote{\url{https://github.com/FayW/DeepLearningTutorials}.}.
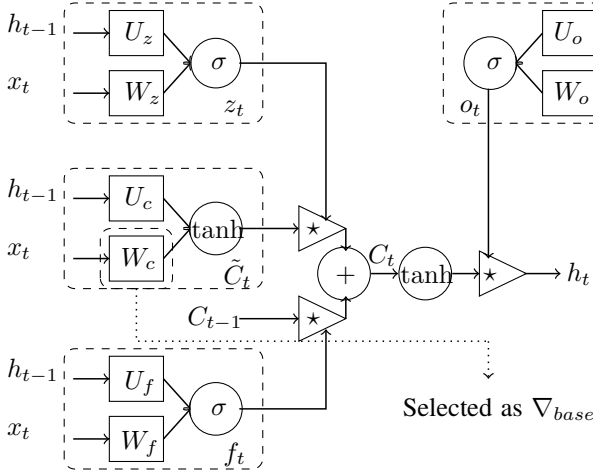
\begin{figure}[t]
	
	\caption{Architecture of Standard LSTM Network}
	\begin{tikzpicture}[x=2.4cm, y=2.0cm]

	\draw (-1.6,-1.35) node[anchor=north]{}
	-- (-1.6,-1.65) node[anchor=north]{}
	-- (-1.34,-1.5) node[anchor=south]{}
	-- cycle;
	\node[text width=2cm] at (-1.14,-1.5) {$\star$};
	
	\draw (-1.6,-1.95) node[anchor=north]{}
	-- (-1.6,-2.25) node[anchor=north]{}
	-- (-1.34,-2.1) node[anchor=south]{}
	-- cycle;
	\node[text width=2cm] at (-1.14,-2.1) {$\star$};

	\draw [line width=0.2mm,->] (-1.34,-2.1)  -- (-1.34,-1.95);
	\draw [line width=0.2mm,->] (-1.34,-1.5)  -- (-1.34,-1.65);
	
	\draw (-1.35,-1.8) circle (0.35cm);
	\node[text width=2cm] at (-0.98,-1.81) {$+$};
	
	\draw (-0.9,-1.8) circle (0.35cm);
	\draw [line width=0.2mm,->] (-1.2,-1.8)   -- (-1.05,-1.8);
	\node[text width=2cm] at (-0.80,-1.68) {$C_{t}$};
	\node[text width=2cm] at (-0.62,-1.8) {$\tanh$};
	\node[text width=2cm] at (-0.16,-1.81) {$\star$};
	\draw [line width=0.2mm,->] (-0.75,-1.8)   -- (-0.6,-1.8);
	\draw [line width=0.2mm,->] (-0.35,-1.8)   -- (-0.15,-1.8);
	\node[text width=2cm] at (0.3,-1.8) {$h_t$};
	\draw (-0.6,-1.65) node[anchor=north]{}
	-- (-0.6,-1.95) node[anchor=north]{}
	-- (-0.34,-1.8) node[anchor=south]{}
	-- cycle;

	\fill [draw,dashed, fill=none,rounded corners] (-2.9,-0.8) rectangle (-1.8,0);
	\fill [draw,fill=none] (-2.65,-0.35) rectangle (-2.35,-0.05);
	\fill [draw,fill=none] (-2.65,-0.75) rectangle (-2.35,-0.45);
	\draw (-2.06,-0.4) circle (0.33cm);
	
	\fill [draw,dashed, fill=none,rounded corners] (-2.9,-1.9) rectangle (-1.8,-1.1);
	\fill [draw,fill=none] (-2.65,-1.45) rectangle (-2.35,-1.15);
	\fill [draw,fill=none] (-2.65,-1.85) rectangle (-2.35,-1.55);
	\draw (-2.06,-1.5) circle (0.35cm);
	
	\fill [draw,dashed, fill=none,rounded corners] (-2.9,-3.1) rectangle (-1.8,-2.3);
	\fill [draw,fill=none] (-2.65,-2.65) rectangle (-2.35,-2.35);
	\fill [draw,fill=none] (-2.65,-3.05) rectangle (-2.35,-2.75);
	\draw (-2.06,-2.7) circle (0.35cm);	
	
	\fill [draw,dashed, fill=none,rounded corners] (-2.7,-1.9) rectangle (-2.3,-1.5);
	\draw [line width=0.2mm,dotted,to path={|- (\tikztotarget)}] (-2.5,-1.9) to (-0.55,-2.25);
	\draw [line width=0.2mm,dotted,->] (-0.55,-2.25) to (-0.55,-2.5);
	\node[text width=3cm] at (-0.4,-2.7) {Selected as $\nabla_{base}$};
	
	\draw [line width=0.2mm,->] (0.25,-0.2)   -- (0.05,-0.2);
	\node[text width=1cm] at (0.29,-0.12) {$h_{t-1}$};
	\draw [line width=0.2mm,->] (0.25,-0.6)   -- (0.05,-0.6);
	\node[text width=1cm] at (0.31,-0.52) {$x_{t}$};
	\node[text width=1cm] at (0.02,-0.2) {$U_{o}$};
	\node[text width=1cm] at (0.02,-0.6) {$W_{o}$};
	\node[text width=1cm] at (-0.35,-0.4) {$\sigma$};
	\node[text width=1cm] at (-0.5,-0.7) {$o_t$};
	
	\draw [line width=0.2mm,->] (-0.25,-0.2) to (-0.4,-0.4);
	\draw [line width=0.2mm,->] (-0.25,-0.6) to (-0.4,-0.4);
	\draw [line width=0.2mm,->] (-0.55,-0.57) to (-0.55,-1.7);
	
	\fill [draw,dashed, fill=none,rounded corners] (-0.8,-0.8) rectangle (0.3,0);
	\fill [draw,fill=none] (-0.25,-0.35) rectangle (0.05,-0.05);
	\fill [draw,fill=none] (-0.25,-0.75) rectangle (0.05,-0.45);
	\draw (-0.54,-0.4) circle (0.35cm);

	\node[text width=2cm] at (-2.8,-0.15) {$h_{t-1}$};
	\draw [line width=0.2mm,->] (-2.85,-0.2)   -- (-2.65,-0.2);
	
	\node[text width=2cm] at (-2.8,-0.55) {$x_{t}$};
	\draw [line width=0.2mm,->] (-2.85,-0.6)   -- (-2.65,-0.6);
	
	\node[text width=2cm] at (-2.8,-1.25) {$h_{t-1}$};
	\draw [line width=0.2mm,->] (-2.85,-1.3)   -- (-2.65,-1.3);
	
	\node[text width=2cm] at (-2.8,-1.65) {$x_{t}$};
	\draw [line width=0.2mm,->] (-2.85,-1.7)   -- (-2.65,-1.7);
	
	\node[text width=2cm] at (-2.8,-2.45) {$h_{t-1}$};
	\draw [line width=0.2mm,->] (-2.85,-2.5)   -- (-2.65,-2.5);
	
	\node[text width=2cm] at (-2.8,-2.85) {$x_{t}$};
	\draw [line width=0.2mm,->] (-2.85,-2.9)  -- (-2.65,-2.9);
	
	\draw [line width=0.2mm,->] (-2.35,-0.2)  -- (-2.2,-0.4);
	\draw [line width=0.2mm,->] (-2.35,-0.6)  -- (-2.2,-0.4);
	
	\draw [line width=0.2mm,->] (-2.35,-1.3)  -- (-2.2,-1.5);
	\draw [line width=0.2mm,->] (-2.35,-1.7)  -- (-2.2,-1.5);
	
	\draw [line width=0.2mm,->] (-2.35,-2.5)  -- (-2.2,-2.7);
	\draw [line width=0.2mm,->] (-2.35,-2.9)  -- (-2.2,-2.7);	
	
	\draw [line width=0.2mm,->,to path={-| (\tikztotarget)}] (-1.93,-0.4) to (-1.45,-1.45);
	\draw [line width=0.2mm,->] (-1.93,-1.5) to (-1.6,-1.5);
	\draw [line width=0.2mm,->] (-1.93,-2.1) to (-1.6,-2.1);
	\draw [line width=0.2mm,->,to path={-| (\tikztotarget)}] (-1.93,-2.7) to (-1.45,-2.17);
	
	\node[text width=2cm] at (-2.15,-0.2) {$U_{z}$};	
	\node[text width=2cm] at (-2.15,-0.6) {$W_{z}$};
	\node[text width=2cm] at (-2.15,-1.3) {$U_{c}$};	
	\node[text width=2cm] at (-2.17,-1.7) {$W_{c}$};	
	\node[text width=2cm] at (-2.15,-2.53) {$U_{f}$};	
	\node[text width=2cm] at (-2.17,-2.93) {$W_{f}$};
	
	\node[text width=2cm] at (-1.67,-0.4) {$\sigma$};
	\node[text width=2cm] at (-1.6,-0.7) {$z_t$};
	\node[text width=0.6cm] at (-2.07,-1.5) {$\tanh$};
	\node[text width=2cm] at (-1.6,-1.8) {$\tilde{C}_t$};
	\node[text width=2cm] at (-1.8,-2.1) {$C_{t-1}$};
	\node[text width=2cm] at (-1.67,-2.7) {$\sigma$};
	\node[text width=2cm] at (-1.6,-3) {$f_t$};
	\end{tikzpicture}
	\label{lstm-fig}
\end{figure}
\subsection{LSTM for Sentence Classification}
LSTM \cite{gers1999learning, graves2012supervised, hochreiter1997long} is an important variant of RNN which solves the gradient-vanishing/explosion problem to a large extent and is widely used in sequence analysis (classification, prediction, etc.) tasks. Our evaluation case is for the sentence analysis which reads in sentences and predict whether it is a negative or positive comment for a movie.
\begin{itemize}
\item Dataset: 25,000 sentences for training set and 1998 for test set.\footnote{Original code uses 1,998 sentences for training and 25,000 for validation. We reverse this for better evaluation.}
\item Objective function: Multiclass cross-entropy.
\item Evaluation metrics: Error rate.
\end{itemize}
\subsubsection{Base-Gradient Selection}
To see how IS is implemented in this variant of RNN, it is necessary to have a look at the detailed training step:
\begin{equation}
\begin{aligned}
&z_t=\sigma(W^zx_{i}^t+U^zh_{t-1}+b^z)\\
&f_t=\sigma(W^fx_{i}^t+U^fh_{t-1}+b^f)\\
&\tilde{C_t}=\tanh(W^cx_{i}^t+U^ch_{t-1}+b^c)\\
&C_t=z_t\ast\tilde{C_t}+f_t\ast C_{t-1}\\
&o_t=\sigma(W^ox_{i}^t+U^oh_{t-1}+b^o)\\
&h_t=o_t\ast\tanh(C_t)
\end{aligned}
\end{equation}
where $x_{i_t}$ is the $t$-th word of training sentence $x_i$, $z_t$ is the input gate, $\tilde{C_t}$ is the buffered memory cell, $f_t$ is the forget gate, $C_t$ is the output memory cell, $o_t$ is the exposure gate and $h_t$ is the output hidden state. See Figure ~\ref{lstm-fig} for illustration.

It can be seen that there are 3 classes of parameters to be learned, i.e., matrices that condition the input $x_{i_t}$, $W^z$, $W^c$, $W^f$, $W^o$, matrices that condition the output state $h_{t-1}$, $U^z$, $U^c$, $U^f$, $U^o$ and bias terms $b^z$, $b^c$, $b^f$, $b^o$. Different from base RNN as we discussed above, in LSTM the output state $h_t$ is further gated by the memory cell $C_t$. 

As we have discussed in section 4, we would like to choose the base-gradient from matrices that conditions $x_i$ rather than $h_i$. We consider $W^c$ as a more important parameters that affects $\tilde{C_t}$ which has the major impact on the output hidden state $h_t$ and thus we choose $\frac{\partial \mathcal{L}_i^T}{\partial W^{c}}$ as our base-gradient which can be derived as:
\begin{equation}
\frac{\partial \mathcal{L}_i^T}{\partial W^{c}}=\sum_{t=1}^{T}\frac{\partial \mathcal{L}_i^t}{\partial y_{t}}\frac{\partial y_t}{\partial h_{t}}\frac{\partial h_t}{\partial C_{t}}\frac{\partial C_t}{\partial \tilde C_{t}}\frac{\partial \tilde C_t}{\partial W^{c}}
\end{equation}
According to our proposed algorithm, we train each sample $x_i$ separately in parallel to obtain its private LSTM model and retrieve its corresponding $\|W_{i}^{c}\|_2$. And thus the sampling distribution $P=\{p_i\}, \forall i\in \{1,2,...,N\}$ is calculated as:
\begin{equation}
	p_i=\frac{\|W_{i}^c\|_2}{\sum_{j=1}^{N}\|W_{j}^c\|_2}, \quad \forall i \in \{1,2,...,N\}
\end{equation}
With a pre-constructed $P$, we generate the training sequence $S$ and adjust step-size with $(Np_i)^{-1}$ correspondingly. 
\begin{figure}[t]
	\centering
	\includegraphics[scale=0.4]{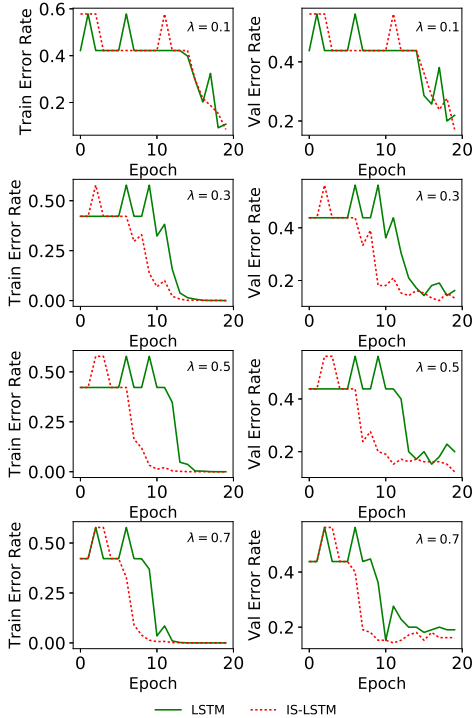}
	\caption{Results of Convergence Acceleration of IS for LSTM. We compare the epoch error rate (calculated at the end of each epoch) between LSTM and IS accelerated LSTM with four different step sizes.}
	\label{lstm_eval}
\end{figure}
\subsubsection{Convergence Acceleration Results of IS-LSTM}
Denote by IS-LSTM as the proposed IS accelerated RNN algorithm, the curves of the error rates of training set and validation set are shown in Figure \ref{lstm_eval}. As can be seen that in all step-size settings, the error rate of IS-LSTM of both training set and validation set drops much faster than traditional LSTM. Particularly, IS-LSTM gains most significant convergence result improvement when $\lambda=0.5$ and decreases when $\lambda$ getting smaller or higher. When step-size is small, e.g., $\lambda=0.1$, the convergence improvement is small. It is reasonable to conclude that there should be an optimal step-size with which IS achieves its maximum performance in accelerating the convergence rate for LSTM. Finding the optimal $\lambda$ for a given LSTM network is a complex problem and we leave this research to future work.

Another noticeable difference is that the error rate of IS-LSTM suffers less variance than LSTM. As can be seen that in the last three rows, significant increase of the error rate happens only once for IS-LSTM while twice for LSTM. 
\begin{figure}[t]
	\caption{Architecture of RNN-RBM}
	\begin{tikzpicture}[x=2.4cm, y=2.0cm]

	\draw [line width=0.2mm,->] (-2.0,-0.6)  -- (-1.5,-0.6);
	\draw [line width=0.2mm,->] (-1.2,-0.6)  -- (-0.7,-0.6);
	\draw [line width=0.2mm,->] (-0.1,-0.6)  -- (0.1,-0.6);
	
	\draw [line width=0.2mm,->] (-1.35,-0.1)  -- (-1.35,-0.45);
	\draw [line width=0.2mm,<->] (-1.35,0.2)  -- (-1.35,0.7);
	
	\draw [line width=0.2mm,->] (0.25,-0.1)  -- (0.25,-0.45);
	\draw [line width=0.2mm,<->] (0.25,0.2)  -- (0.25,0.7);
	
	\draw [line width=0.2mm,->] (-0.55,-0.1)  -- (-0.55,-0.45);
	\draw [line width=0.2mm,<->] (-0.55,0.2)  -- (-0.55,0.7);

	\fill [draw,fill=none] (0.1,0.7) rectangle (0.4,1);
	\fill [draw,fill=none] (-1.5,0.7) rectangle (-1.2,1);
	\fill [draw,fill=none] (-0.7,0.7) rectangle (-0.4,1);
	
	\fill [draw,fill=none] (0.1,-0.1) rectangle (0.4,0.2);
	\fill [draw,fill=none] (-1.5,-0.1) rectangle (-1.2,0.2);
	\fill [draw,fill=none] (-0.7,-0.1) rectangle (-0.4,0.2);
	
	\fill [draw,fill=none] (-2.3,-0.75) rectangle (-2.0,-0.45);
	\fill [draw,fill=none] (-1.5,-0.75) rectangle (-1.2,-0.45);
	\fill [draw,fill=none] (-0.7,-0.75) rectangle (-0.4,-0.45);
	\fill [draw,fill=none] (0.1,-0.75) rectangle (0.4,-0.45);
	\node[text width=2cm] at (0.6,-0.6) {$u_T$};
	\node[text width=2cm] at (-0.2,-0.6) {$u_2$};
	\node[text width=2cm] at (-1,-0.6) {$u_1$};
	\node[text width=2cm] at (-1.8,-0.6) {$u_0$};
	\node[text width=2cm] at (-1,0.85) {$h_1$};
	\node[text width=2cm] at (-0.2,0.85) {$h_2$};
	\node[text width=2cm] at (0.6,0.05) {$h_T$};
	\node[text width=2cm] at (-1,0.05) {$v_1$};
	\node[text width=2cm] at (-0.2,0.05) {$v_2$};
	\node[text width=2cm] at (0.6,0.85) {$v_T$};
	\node[text width=2cm] at (0.1,0.85) {$\boldsymbol{\cdot}\boldsymbol{\cdot}\boldsymbol{\cdot}$};
	\node[text width=2cm] at (0.1,0.05) {$\boldsymbol{\cdot}\boldsymbol{\cdot}\boldsymbol{\cdot}$};
	\node[text width=2cm] at (0.1,-0.61) {$\boldsymbol{\cdot}\boldsymbol{\cdot}\boldsymbol{\cdot}$};
	\node[text width=2cm] at (-1.25,0.62) {$b^h_1$};
	\node[text width=2cm] at (-0.45,0.62) {$b^h_2$};
	\node[text width=2cm] at (0.35,0.62) {$b^h_T$};

	\node[text width=2cm] at (-1.23,0.02) {$b^v_1$};
	\node[text width=2cm] at (-0.43,0.02) {$b^v_2$};
	\node[text width=2cm] at (0.32,0.02) {$b^v_T$};
	\node[text width=2cm] at (-1.8,0.13) {$W^{uh}$};
	\node[text width=2cm] at (-1.53,-0.14) {$W^{uv}$};
	\node[text width=2cm] at (-1.45,-0.5) {$W^{uu}$};
	\node[text width=2cm] at (-1.25,-0.3) {$W^{vu}$};
	\node[text width=2cm] at (-0.85,0.4) {$W$};
	\fill [draw,dashed, fill=none,rounded corners] (-1.31,0.25) rectangle (-1.05,0.55);
	\draw [line width=0.3mm,dotted,to path={-| (\tikztotarget)}] (-1.05,0.4) to (0.65,0.4);
	\draw [line width=0.3mm,dotted,->] (0.65,0.4)  -- (0.65,-0.05);
	\node[text width=1.5cm] at (0.75,-0.3) {Selected as $\nabla_{base}$};

	\draw [line width=0.2mm,->] (-2.2,-0.45)  -- (-1.65,0.55);
	\draw [line width=0.2mm,->] (-2.0,-0.45)  -- (-1.5,-0.1);
	\draw [line width=0.2mm,->] (-1.3,-0.45)  -- (-0.85,0.55);
	\draw [line width=0.2mm,->] (-1.2,-0.45)  -- (-0.7,-0.1);
	\draw [line width=0.2mm,->] (-0.3,0.15)  -- (-0.05,0.5);
	\draw [line width=0.2mm,->] (-0.3,-0.45)  -- (-0.05,-0.1);
	\end{tikzpicture}
	\label{rnnrbm-fig}
\end{figure}
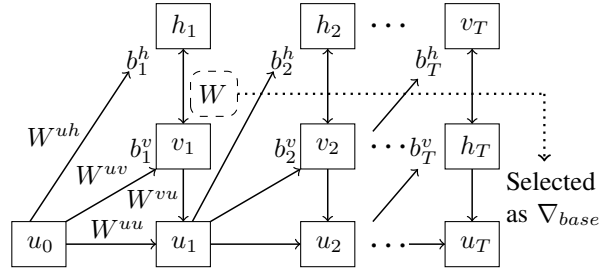
\subsection{RNN-RBM for Polyphonic Music Modeling}
Besides of NLP related tasks, acoustic sequence processing is another important application field for RNNs. We thus choose RNN-based Restricted Boltzmann Machines (RNN-RBM) polyphonic music modeling \cite{boulanger2012modeling} as our second evaluation case of IS for RNN. Particularly, in this case RNN is used to condition a specified modeling structure RBM\cite{hinton2010practical} which can be seen as another variant of RNN. It is every interesting to observe how IS performs in this kind of multilayer-RNN architectures. 
\begin{itemize}
	\item Dataset: 147,737 musical sequences with dimensions as 88.
	\item Objective function: Cross-entropy.
	\item Evaluation metrics: Object function value.
\end{itemize}
\subsubsection{Restricted Boltzmann Machines}
RBM models complicate distributions based on its energy function (which is to be learned during training). Denote $W$ ($W^T$ as its transpose) represents the weights connecting hidden units $h$ and visible units $v$, and $b^v$, $b^h$ are the offsets of the visible and hidden layers respectively. The update rule of the parameters are defined as:
\begin{equation}
\begin{aligned}
&h_{t+1}\sim \sigma(W^Tv_t+b^h_t)\\
&v_{t+1}\sim \sigma(Wh_{t+1}+b^v_t)\\
\end{aligned}
\label{rbm}
\end{equation}
where $\sigma$ is typically set as sigmoid function. The above equations mean that $h_{t+1}$ is activated with probability $\sigma(W^Tv_t+b^h_t)$ and similar for $v_{t+1}$.
\subsubsection{RNN-RBM}
RNN extends the ability of RBM which makes it able to model multi-modal conditional distributions. With the combination of RNN, an extra recurrent procedure is added as:
\begin{equation}
\begin{aligned}
\label{rnnrbm}
\centering
&b^v_t=b^v+W^{uv}u_{t-1}\\
&b^h_t=b^h+W^{uh}u_{t-1}\\
&u_t=\tanh(b^u+W^{uu}u_{t-1}+W^{vu}v_t)
\end{aligned}
\end{equation}
where $v_t$ is the input vector, $u_t$ is the output hidden state of the added RNN and $b^v_t$, $b^h_t$ are intermediate results to be calculated. And $b^v, b^h, b^u$, $W^{uv}$, $W^{uh}$, $W^{uu}$, $W^{vu}$ are parameters to be learned. With the calculated $b^v_t$ and $b^h_t$, $v$ serves as the initial input of Equation \ref{rbm}, i.e., $v_0$. The unrolled architecture of RNN-RBM is shown in Figure ~\ref{rnnrbm-fig}. 
\subsubsection{Base-Gradient Selection}
From an architectural point of view, RNN-RBM is actually a two layer RNNs with the RBM (seen as a variant of RNN) as the second layer where the output is generated. In considering the fact that the first layer RNN is for the calculation of auxiliary parameters $b^t$ and $b^v$ as the parameter of the second RNN, we choose $W$ of Equation \ref{rbm} (the RBM recursion) which conditions both input $v_t$ and hidden state $h_t$ as the base-gradient since we consider it serves similar function with $W^x$ in base RNN which is a proper proxy of $\nabla_{base}$. Accordingly, we have $p_i=\frac{\|W_i\|_2}{\sum_{j=1}^{N}\|W_j\|_2}, \forall i \in \{1,2,...,N\}$.
\begin{figure}[t]
	\centering
	\includegraphics[scale=0.4]{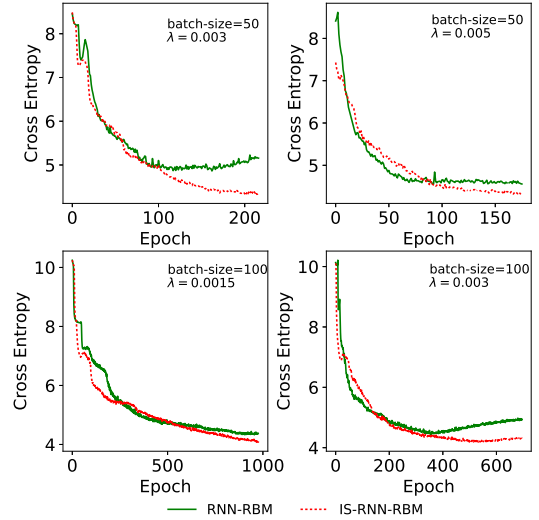}
	\caption{Convergence Results. We compare the epoch objective function between standard non-IS RNN-RBM and IS accelerated RNN-RBM with two different batch-sizes.}
	\label{rbm_eval}
\end{figure}
\subsubsection{Convergence Acceleration Results of IS-RNN-RBM}
We use two different batch-sizes to comply with the original version of this case\footnote{In this case, the batch-size actually indicates how many consecutive musical slices are combined as one training sample.}. We evaluate two different step-sizes for each batch-size, Figure ~\ref{rbm_eval} shows the curves of cost per epoch\footnote{By summing the objective cost for each sample (without model update) and then return its average at the end of each epoch.}. As can be seen that IS accelerated RNN-RBM achieves better convergence results than standard RNN-RBM in both batch-sizes. In the first configuration, i.e. batch-size $=50$ and $\lambda=0.3$, the objective cost of standard RNN-RBM encounters heavy regression when the cost is low while for IS accelerated RNN-RBM, the objective cost continues to decrease which is a very significant improvement. Similar result can also be observed when batch-szie $=100$ and $\lambda=0.003$. 

This clearly shows the effect of IS, it makes the stochastic training procedure more robust since its searching direction is statistically better (by choosing important data) than Non-IS trainings. Such robustness does not only benefit the convergence rate, but also very helpful in avoiding accuracy regression that frequently happens when step-size is relative large (as shown in this case). In fact, this actually means that IS accelerated RNN can use larger step-sizes and less-likely to be trapped in local-minimum while non-IS RNN is not able to. This surely leads to a faster convergence rate and sometimes higher final accuracy.
\subsection{Remark}
Although we have analyzed the expected attributes of proper base-gradient in theoretical, its selection is still somehow empirical. In general, the matrix that conditions the (embedded) input performs the best. Another important experience is that the accuracy of the FIM algorithm, i.e., $\epsilon$ in Algorithm 2 has significant effect on the performance of IS. By setting a higher accuracy, i.e., a lower $\epsilon$, the total iterations $T_i$ needed by each separate fast approximation training thread varies largely which potentially incurs larger differences between the trained models and consequently the target gradients. Empirically, we consider this will reflect the relative importance of data sample for training more significantly and benefits the performance of IS.
\section{Conclusion}
Due to the practical significance and novelty, we are motivated to apply IS in RNN for convergence acceleration. The calculation of the sampling distribution $P$ w.r.t which the IS is based on requires the training data to be well-structured. However for RNNs the input data are often randomly mapped before training which is the major bottlenecking problem that prevents the effective application of IS in RNNs. To break this obstacle, we propose an optimized IS procedure based on Fast-Importance-Mining (FIM) algorithm which trains each input data $x_i, \forall i\in\{1,2,...,N\}$ separately with its private model until certain convergence accuracy $\epsilon$ is met. IS then use the selected base-gradient's norm $\|\nabla_{base}\|_2$ as the proxy value of the sampling importance upon on which the sampling distribution $P$ can be constructed.

We evaluate our optimized IS accelerated RNN on two popular applications. In both cases, we the select $\nabla_{base}$ accordingly and the results show that IS accelerated RNN-based optimizations achieves better convergence results (convergence rate or final accuracy) than its non-IS counterparts. We also notice that certain relationship exists between convergence improvements and step-size, batch-size, etc. In practical, FIM incurs small additional time cost since the algorithm can be totally parallelized and the training for single $x_i$ is fast. Related source code of FIM and the evaluation applications are all accessible on the author's github repository.
\section*{Acknowledgements}
The authors thank Jason Ye, Professor Guihai Chen and Professor Xiaofeng Gao for their important helps.
\bibliography{final}

\begin{thebibliography}{10}

\bibitem{alain2015variance}
{\sc G.~Alain, A.~Lamb, C.~Sankar, A.~Courville, and Y.~Bengio}, {\em Variance
  reduction in sgd by distributed importance sampling}, arXiv preprint
  arXiv:1511.06481,  (2015).

\bibitem{Alain}
{\sc L.~A. S. C. C.~A. Alain, Guillaume and Y.~Bengio}, {\em Variance reduction
  in sgd by distributed importance sampling.}, in arXiv preprint
  arXiv:1511.06481., 2015.

\bibitem{bergstra2010theano}
{\sc J.~Bergstra, O.~Breuleux, F.~Bastien, P.~Lamblin, R.~Pascanu,
  G.~Desjardins, J.~Turian, D.~Warde-Farley, and Y.~Bengio}, {\em Theano: A cpu
  and gpu math compiler in python}, in Proc. 9th Python in Science Conf, 2010,
  pp.~1--7.

\bibitem{boulanger2012modeling}
{\sc N.~Boulanger-Lewandowski, Y.~Bengio, and P.~Vincent}, {\em Modeling
  temporal dependencies in high-dimensional sequences: Application to
  polyphonic music generation and transcription}, arXiv preprint
  arXiv:1206.6392,  (2012).

\bibitem{csiba2016importance}
{\sc D.~Csiba and P.~Richt{\'a}rik}, {\em Importance sampling for minibatches},
  arXiv preprint arXiv:1602.02283,  (2016).

\bibitem{funahashi1993approximation}
{\sc K.-i. Funahashi and Y.~Nakamura}, {\em Approximation of dynamical systems
  by continuous time recurrent neural networks}, Neural networks, 6 (1993),
  pp.~801--806.

\bibitem{gers1999learning}
{\sc F.~A. Gers, J.~Schmidhuber, and F.~Cummins}, {\em Learning to forget:
  Continual prediction with lstm},  (1999).

\bibitem{Siddharth}
{\sc S.~Gopal.}, {\em Adaptive sampling for sgd by exploiting side
  information.}, in Proceedings of the 33 rd International Conference on
  Machine Learning., 2016.

\bibitem{graves2012supervised}
{\sc A.~Graves et~al.}, {\em Supervised sequence labelling with recurrent
  neural networks}, vol.~385, Springer, 2012.

\bibitem{hazan2011beating}
{\sc E.~Hazan, T.~Koren, and N.~Srebro}, {\em Beating sgd: Learning svms in
  sublinear time}, in Advances in Neural Information Processing Systems, 2011,
  pp.~1233--1241.

\bibitem{hinton2010practical}
{\sc G.~Hinton}, {\em A practical guide to training restricted boltzmann
  machines}, Momentum, 9 (2010), p.~926.

\bibitem{hochreiter1997long}
{\sc S.~Hochreiter and J.~Schmidhuber}, {\em Long short-term memory}, Neural
  computation, 9 (1997), pp.~1735--1780.

\bibitem{mikolov2010recurrent}
{\sc T.~Mikolov, M.~Karafi{\'a}t, L.~Burget, J.~Cernock{\`y}, and
  S.~Khudanpur}, {\em Recurrent neural network based language model.}, in
  Interspeech, vol.~2, 2010, p.~3.

\bibitem{Bach}
{\sc E.~Moulines and F.~R. Bach}, {\em Non-asymptotic analysis of stochastic
  approximation algorithms for machine learning}, in Advances in Neural
  Information Processing Systems, Curran Associates, Inc., 2011, pp.~451--459.

\bibitem{Needell}
{\sc D.~Needell, R.~Ward, and N.~Srebro}, {\em Stochastic gradient descent,
  weighted sampling, and the randomized kaczmarz algorithm}, in Advances in
  Neural Information Processing Systems, Curran Associates, Inc., 2014,
  pp.~1017--1025.

\bibitem{pineda1987generalization}
{\sc F.~J. Pineda}, {\em Generalization of back-propagation to recurrent neural
  networks}, Physical review letters, 59 (1987), p.~2229.

\bibitem{Strohmer}
{\sc T.Strohmer and R.~Vershynin.}, {\em A randomized kaczmarz algorithm with
  exponential convergence}, in The Journal of Fourier Analysis and
  Applications., vol.~2, 2009, pp.~262--278.

\bibitem{wang1993analysis}
{\sc J.~Wang}, {\em Analysis and design of a recurrent neural network for
  linear programming}, IEEE Transactions on Circuits and Systems I: Fundamental
  Theory and Applications, 40 (1993), pp.~613--618.

\bibitem{p_zhao}
{\sc P.~Zhao and T.~Zhang.}, {\em Stochastic optimization with importance
  sampling for regularized loss minimization.}, in Proceedings of the 32nd
  International Conference on Machine Learning., 2015.

\end{thebibliography}
\bibliographystyle{siamplain}
\end{document}